\documentclass[]{article}
\usepackage[margin=0.75in]{geometry}
\usepackage{libertine}

\usepackage{graphics} 
\usepackage{epsfig} 
\usepackage{amsmath} 
\usepackage{amssymb}  
\usepackage{bm}
\usepackage{subfig}
\usepackage{array}
\usepackage{algorithm, algorithmic}

\title{\LARGE \bf
Safe rendering of high impedance on a series-elastic actuator with disturbance observer-based torque control
}
\author{Kevin Haninger$^1$, Abner Asignacion$^2$, Sehoon Oh$^2$\footnote{This paper submitted to ICRA 2020 \newline $^1$ Affiliated with the Division of Automation Technology, Fraunhofer Institute for Production Systems and Design Technology (IPK), Berlin, Germany (\texttt{kevin.haninger@ipk.fraunhofer.de}) \newline $^2$ Affiliated with the Department of Robotic Engineering, Daegu Gyeongbuk Institute of Science and Technology, Daegu, Korea}}
\date{}

\begin{document}
	
\maketitle  

\begin{abstract}
	An important performance metric for series-elastic actuators is the range of impedance which they can safely render. Advanced torque control, using techniques such as the disturbance observer, improve torque tracking bandwidth and accuracy, but their impact on safe impedance range is not established.  However, to define a safe impedance range requires a practical coupled stability condition.  Here, passivity-based conditions are proposed for two variants of DOB torque control, and validated experimentally in a high-stiffness environment. While high-gain PD torque control has been shown to reduce Z-width, it is here shown that a DOB reduces the need for high-gain PD feedback and allows a higher rendered impedance.  A dynamic feedforward compensator is proposed which increases the maximum safe impedance of the DOB, validated in experimentally in collision with high-stiffness environments and manual excitation.
\end{abstract}

\section{Introduction}
Torque control of series-elastic actuators (SEAs) \cite{pratt1995} brings them closer to ideal torque sources, supporting higher-level control such as impedance control \cite{hogan1984}.  Torque control performance, in the sense of torque tracking bandwidth and accuracy, can be improved with high-gain feedback control, realized with PID compensators or the disturbance observer (DOB) \cite{ohnishi1996,kong2009a}. However, the torque control should also support other senses of performance, such as allowing the safe rendering of a wide range of impedance. In particular, rendering a high impedance offers potential benefits, such as improving traditional motion control performance while maintaining a low high-frequency impedance, reducing collision force \cite{albu-schaffer2008, haninger2019}.

Recent work has suggested that high-gain torque control reduces the Z-width of an actuator (the range of impedance parameters which are `achievable' \cite{colgate1994}) on hydraulic \cite{boaventura2013} and electric actuators \cite{focchi2016}. However, the impedance parameter may not correspond to the rendered stiffness, especially if the torque control has low torque-tracking performance. An alternative performance metric which addresses this potential gap is Z-region \cite{zhao2018}, which considers the range of impedance which can be rendered at the interaction port. However, all of these dynamic range metrics require a realistic means of finding safe or realizable maximum/minimum impedance parameters.  This safety range can be found experimentally \cite{colgate1994}, but for design, a practical model-based condition for evaluating safety is needed.  

The classical formalization of safety for interactive robots is passivity,  which allows coupled stability with an arbitrary passive environment. This has been used as a design condition for hierarchical control of SEAs with: torque/velocity control \cite{vallery2007}, impedance/torque/velocity control \cite{tagliamonte2014b,tosun2019}, and in comparative studies \cite{mosadeghzad2012, calanca2017}. Analysis based on passivity has given insight into the importance of motor damping as well as torque-control derivative action  \cite{colgate1997, tagliamonte2014b}. However, passivity conditions for DOB torque control, with or without an impedance controller, are not established.  Many of the stability proofs for DOB torque control consider environments with bounded parametric uncertainty, raising questions about stability in arbitrary environments.

\begin{figure}[t]
	\centering
	\includegraphics[width=.5\columnwidth]{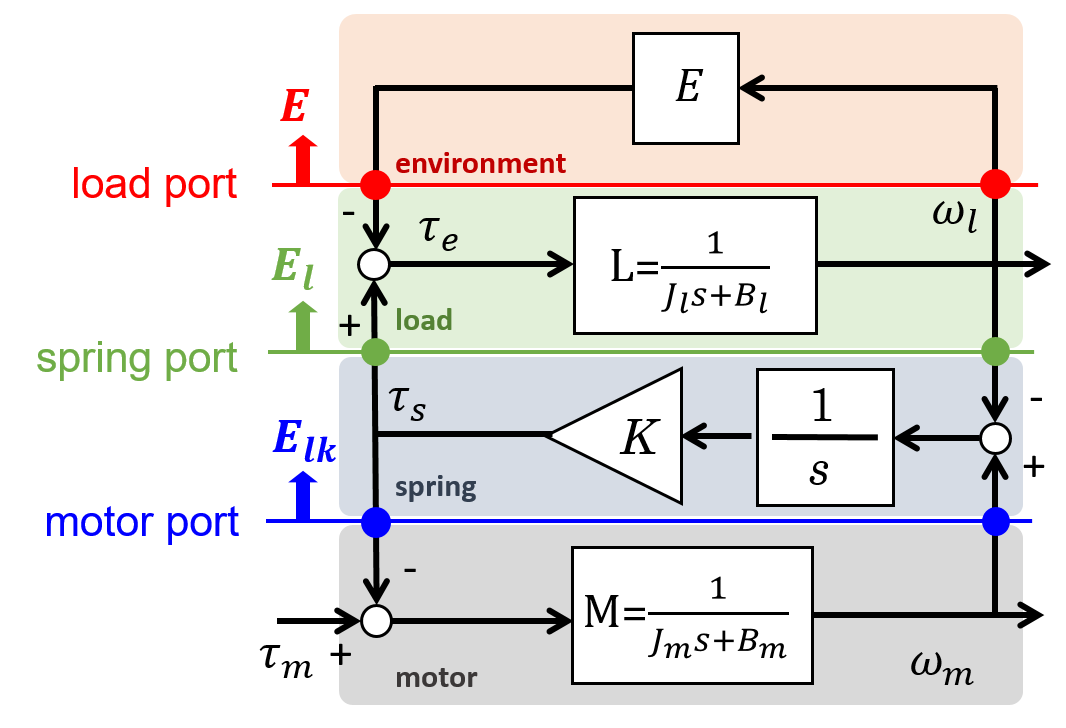}
	\caption{Schematic for an SEA integrated to an application, with motor dynamics $M$ coupled to load dynamics $L$ through the elastic element $K$. The environment $E$ couples to the load dynamics. Three power-continuous interaction ports are shown.\label{port_definitions}} 
\end{figure}

To be practical for design, the safety condition should not be too conservative. As the conservatism of passivity is well-known \cite{buerger2007, mehling2014}, various passivity relaxations have been proposed, including analyzing passivity at different ports \cite{lee2019}, and replacing passivity with a mixed passivity/small-gain condition \cite{haninger2018a}. Passivity observers have also been implemented on SEAs \cite{losey2016, lee2019}, enforcing passivity by modifying the control signal, at the cost of complexity and performance.

The DOB is used in many state-of-the-art SEA controllers, including rotary \cite{kong2009a,mehling2015,oh2017}, linear \cite{paine2014}, cable-driven \cite{lu2015}, and multiple-stiffness \cite{sariyildiz2015a} SEAs. Performance is typically shown by the bandwidth and accuracy (disturbance rejection, reference tracking) of the torque loop. Outer-loop position control has been investigated with a DOB \cite{paine2014, sariyildiz2015a}. Impedance control with inner-loop DOB is also established \cite{mehling2015}, and the DOB is shown to reduce the need for high-gain torque feedback. However, the ability to render high impedances with a DOB, especially in high-stiffness environments, is not yet investigated.

Towards high impedance rendering with DOB torque control, two DOB architectures are considered in this paper, one closed around the motor dynamics \cite{kong2009a}, and one around the torque dynamics \cite{oh2017}, referred to as DOBm and DOBt respectively. To realistically evaluate the safe impedance range, this paper first develops and validates coupled stability conditions based on passivity, using load dynamics to relax conservatism. These conditions motivate a realistic evaluation of the Z-region and maximum rendered stiffness, allowing analysis of the impact of the DOB and the design of feedforward/feedback compensators for a large Z-region.  A dynamic feedforward compensator is proposed which increase the upper impedance limit for both DOB architectures. Experimental validation demonstrates the coupled stability conditions, as well as compares maximum impedance rendering among the controller architectures. 

\section{Control Architecture and Metrics \label{safety_and_performance}}
This section introduces DOB control for impedance control, as well as safety and performance metrics which will be considered.
\begin{figure}
\centering
\subfloat[]{\includegraphics[width=.5\columnwidth]{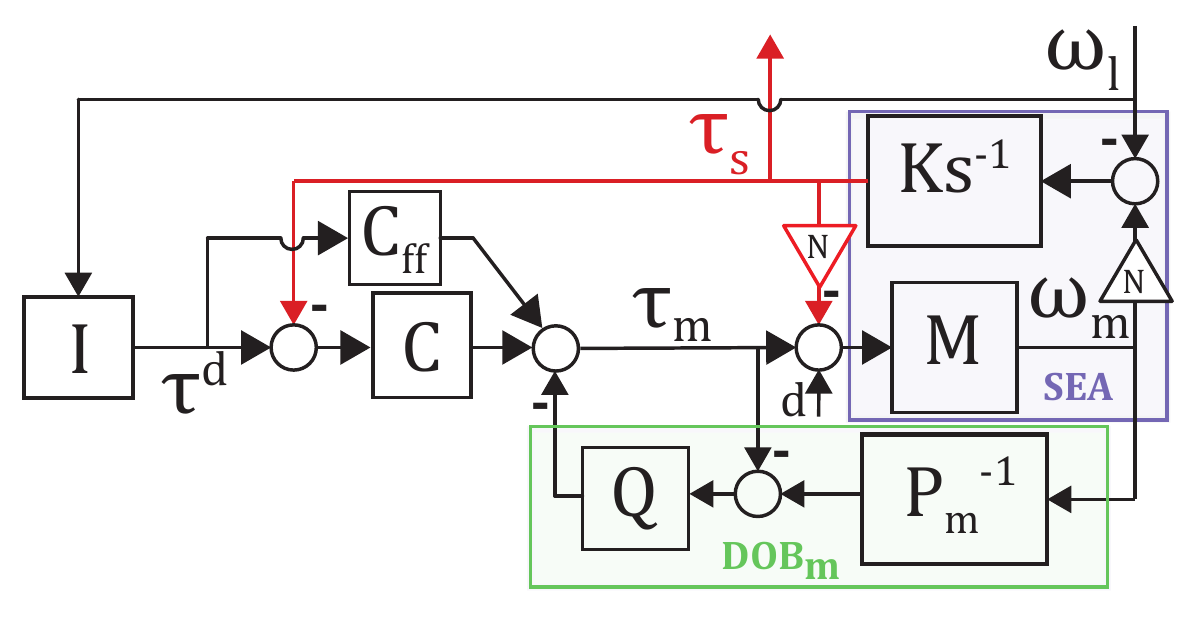}} 
\subfloat[]{\includegraphics[width=.5\columnwidth]{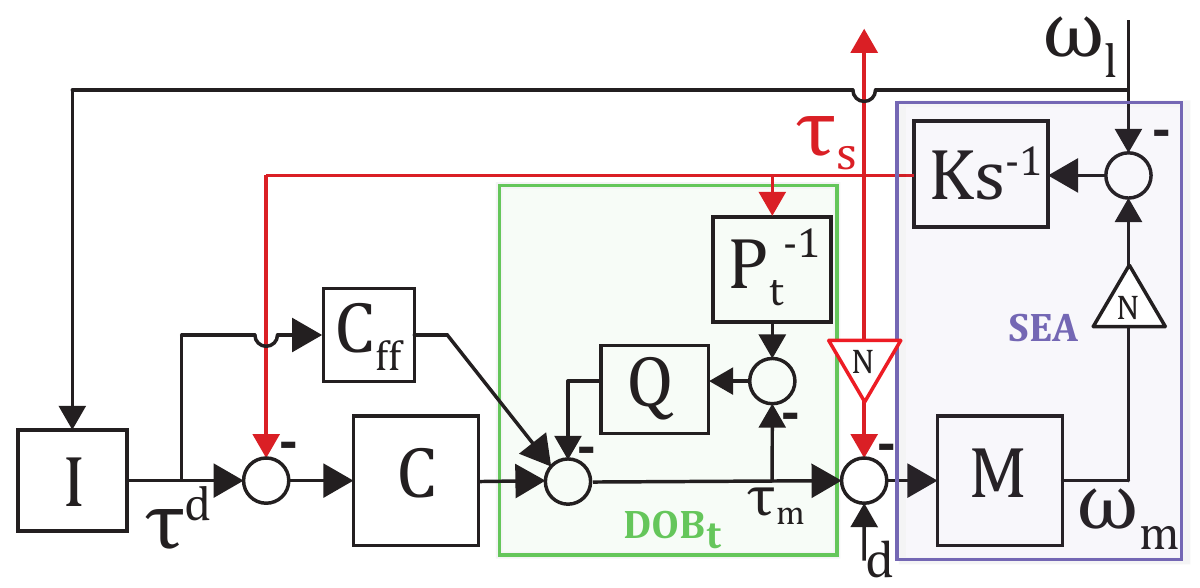}}
\caption{Two DOB architectures for SEAs, where (a) uses the DOB to enforce the model motor dynamics, and (b) directly uses the DOB to enforce torque dynamics. The inverse of nominal dynamics $P_{m}$ and $P_t$ are used, along with tuning filter $Q$. Feedback controller $C$ and feedforward control $C_{ff}$ are included. Inverse gear ratio $N$ relates motor dynamics $M$ to the spring $K$, and disturbance force $d$ describes unmodelled gearbox friction \label{DOB_architecture}}
\end{figure}

\subsection{Disturbance Observer \label{DOB_intro}}
The disturbance observer is a frequency-domain control technique often used in precision motion control \cite{ohnishi1996} which uses the inverse of a nominal plant model to estimate external disturbances, then adds the negative of the estimated disturbance to the control input for cancellation. In addition to the rejection of disturbances, the DOB enforces nominal dynamics, making the plant appear as the nominal model within the $Q$-filter bandwidth, improving the accuracy of feed-forward compensators \cite{yun2012}. 

In applying the DOB to SEAs, the DOB is classically applied around the motor dynamics \cite{kong2009a}, but recent work has shown advantages in applying the DOB to the torque output \cite{oh2017}. Denoting these DOBm and DOBt (respectively), their block diagram can be seen in Figure \ref{DOB_architecture}. The spring port dynamics which result can be seen in Table \ref{port_dyn}.

The inverse models considered here are the nominal models of the effective plant of the DOB loop under fixed output ($P_m=\hat{M}$ and $P_t=\mathtt{fb}(\hat{M},Ks^{-1})$). Second-order low-pass filters are used for $Q$ filters, with cutoff frequency $\omega_{DOB}$. 

\subsection{Coupled stability \label{safety_conditions_section}}
To conclude the coupled stability of the SEA with an arbitrary passive environment, the SEA and environment can be separated at any of the ports shown in Figure \ref{port_definitions}. Often, the passivity of the SEA at the spring port is investigated \cite{vallery2007, tosun2019}, while other work has investigated passivity at the load port \cite{lee2020, lee2019, tagliamonte2014b}. The load dynamics are the mechanism permanently attached to the output of the SEA, which are typically known. Given a controller, the dynamics rendered at the spring port can be seen in Table \ref{port_dyn}, and readily transformed into the load dynamics as
\begin{eqnarray}
\frac{\omega_l}{\tau_e} = \frac{L}{1+LZ_s}
\end{eqnarray}
where $Z_s$ is the spring dynamics from $\omega_l$ to $\tau_s$, as specified in Table \ref{port_dyn}. 

\begin{table*}
	\renewcommand{\arraystretch}{2}
	\centering
	\begin{tabular}{|r|l|l|l|}
		\hline 
		& Spring port dynamics & Z-region & Rendered stiffness\tabularnewline
		\hline 
		\hline 
		DOBm & $\tau_s=\frac{Ks^{\texttt{-}1}\left(N\tilde{M}\left(C\texttt{+}C_{ff}\right)I\texttt{+}1\right)\omega_{l}\texttt{+}Ks^{\texttt{-}1}\tilde{M}\left(1\texttt{-}Q\right)Nd}{1\texttt{+}Ks^{\texttt{-}1}N\tilde{M}\left(C\texttt{+}N(1\texttt{-}Q)\right)}$ & $\int_0^\infty\ln\left|N\tilde{M}\left(C\texttt{+}C_{ff}\right)I_{u}\texttt{+}1\right|d\omega$ & $\tau_s|_{\omega=0}=\frac{\left(K_{p}\texttt{+}C_{ff}\right)K_{imp}\theta_{l}\texttt{+}\left(1\texttt{-}Q\right)d}{\left(K_{p}\texttt{+}1\texttt{-}Q\right)}$\tabularnewline
		\hline 
		DOBt & $\tau_s=\frac{Ks^{\texttt{-}1}\left(NM\left(C\texttt{+}C_{ff}\right)I\texttt{+}1\texttt{-}Q\right)\omega_{l}\texttt{+}Ks^{\texttt{-}1}M\left(1\texttt{-}Q\right)Nd}{1\texttt{-}Q\texttt{+}QMP_{t}^{\texttt{-}1}\texttt{+}Ks^{\texttt{-}1}NM\left(C\texttt{+}N(1\texttt{-}Q)\texttt{+}QMP_{t}^{\texttt{-}1}\right)}$ & $\int_0^\infty\ln\left|N\frac{M\left(C\texttt{+}C_{ff}\right)I_{u}}{1\texttt{-}Q}\texttt{+}1\right|d\omega$ & $\tau_s|_{\omega=0}=\frac{\left(K_{p}\texttt{+}C_{ff}\right)K_{imp}\theta_{l}\texttt{+}\left(1\texttt{-}Q\right)d}{\left(K_{p}\texttt{+}1\texttt{-}Q\texttt{+}QMP_{t}^{\texttt{-}1}\right)}$\tabularnewline
		\hline 
		No DOB & $\tau_s=\frac{Ks^{\texttt{-}1}\left(NM\left(C\texttt{+}C_{ff}\right)I\texttt{+}1\right)\omega_{l}\texttt{+}Ks^{\texttt{-}1}MNd}{1\texttt{+}Ks^{\texttt{-}1}NM\left(C\texttt{+}N\right)}$ & $\int_0^\infty\ln\left|NM\left(C\texttt{+}C_{ff}\right)I_{u}\texttt{+}1\right|d\omega$ & $\tau_s|_{\omega=0}=\frac{\left(K_{p}\texttt{+}C_{ff}\right)K_{imp}\theta_{l}\texttt{+}d}{\left(K_{p}\texttt{+}1\right)}$\tabularnewline
		\hline 
	\end{tabular}
	
	\caption{Spring port dynamics and performance metrics of the three  architectures on a two-inertia SEA model, with dynamics as defined in Figure \ref{DOB_architecture}, with $\tilde{M}= M(1-Q\Delta)^{-1}$, where $M=\hat{M}(1+\Delta)$ and $\theta = \omega/s$.\label{port_dyn}}
\end{table*}

\subsection{Performance metrics}
The typical design metric for torque control is torque tracking bandwidth with fixed output. Although useful for comparison, its generalization to lower-impedance environments and more application-driven requirements (e.g. large Z-region) are unclear. The performance of impedance control can also be characterized by the accuracy of the rendered impedance \cite{haninger2016a}, but in many applications the range of impedance which can be rendered is more important than their accuracy.

\subsubsection{Z-region}
Z-width \cite{colgate1994} characterizes the range of impedance parameters which can be achieved, originally proposed on haptic systems. For SEAs with inner-loop torque control, the impedance parameters do not directly determine rendered dynamics - they depend on the torque control.  To address this, Z-region is proposed \cite{zhao2018}, which measures the range in impedances which can be rendered at the interaction port. Supposing upper and lower impedance limits at a port are known, $Z_u$ and $Z_l$, the metric is
\[
Z_{region}=\int_{\omega_{1}}^{\omega_{2}}W\left(\omega\right)\left|\ln\left|Z_{u}\left(j\omega\right)\right|-\ln\left|Z_{l}\left(j\omega\right)\right|\right|d\omega
\]
Letting $W\left(\omega\right)=1$, the objective can be rewritten as:
\begin{eqnarray}
Z_{region}=\int_{\omega_{1}}^{\omega_{2}}\left|\ln\left|\frac{Z_{u}\left(j\omega\right)}{Z_{l}\left(j\omega\right)}\right|\right|d\omega.
\end{eqnarray}
Here, $Z_{l}$ is assumed to be zero-impedance control (i.e. $I\left(s\right)=0$, which is rarely problematic), which simplifies the metric. Furthermore, at high frequencies where $Z_u=Z_l=Ks^{-1}$, the integrand of Z-region goes to zero, thus the simplification $\omega_2\rightarrow \infty$ is used here. The resulting expressions can be seen in Table \ref{port_dyn}.

\subsubsection{Maximum rendered stiffness}
In addition to Z-region, the highest rendered stiffness is considered.  These expressions can be derived by taking the limit of spring port dynamics as $\omega\rightarrow 0$, and are seen in the right column of Table \ref{port_dyn}.

\subsection{Performance with coupled stability}
To consider the real-world achievable performance, a realistic coupled stability constraint must be considered.  A coupled stability condition is spring port passivity - i.e. the positive-realness of $\omega_l\rightarrow \tau_s = Z_s$. Designing the controllers to maximize safe rendered stiffness is then

\begin{eqnarray}
 \max_{K_{imp}, C, C_{ff}, Q}  \,\,\,\,\lim_{\omega\rightarrow 0}|\omega Z_s(j\omega)| & \mathrm{s.t.} & \mathtt{Re}(Z_s)>0.
\end{eqnarray}

With this safety condition, the maximum rendered stiffness is $K$, the physical stiffness.  As outlined in Section \ref{safety_conditions_section}, the passivity at the load port can be considered.  Inverting the load port dynamics so they are an impedance; only the damping of the load dynamics contribute to the real response. Or the load dynamics can be directly used, allowing either of the following:
\begin{eqnarray}
\max_{K_{imp}, C, C_{ff}, Q}  \lim_{\omega\rightarrow 0}|\omega Z_s| & \mathrm{s.t.} & \mathtt{Re}\{Z_l\}>B_l \\
\max_{K_{imp}, C, C_{ff}, Q}  \lim_{\omega\rightarrow 0}|\omega Z_s| & \mathrm{s.t.} & \mathtt{Re}\{\frac{L}{1+LZ_s}\}>0 \label{load_pass}
\end{eqnarray}

Though complex, the load passivity can be written in closed form (with the aid of a symbolic algebra system), for the no DOB case giving
\begin{eqnarray}
\mathtt{Re}\{Z_l\}&=&\frac{c_4\omega^4+c_2\omega^2+c_0}{KN(J_mK_d\omega^2-KK_dN^2+B_mK_p)}  \\
c_4 &=& B_lJ_m \nonumber\\
c_2 &=& B_lB_m+2B_lB_mKK_dN+J_mK_dK^2N \cdots \nonumber\\
& & -2B_lJ_mKN(K_p+1)+B_lK_d^2K^2N^2 \\
c_0 &=& K^2N(B_mK_p+B_mN+B_lK_p^2N) \cdots \\
& =& +K^2N(2B_lK_pN^2+B_lN^3)
\end{eqnarray}

These expressions can be analyzed to inform the choice of control parameters in the next section. 

\section{Controller Architecture and Design}
This section analyzes and designs controllers with consideration of the safety and performance conditions in Section \ref{safety_and_performance}. Parameter settings for high-impedance rendering are motivated, as well as a dynamic feed-forward compensation which improves the safe high-impedance limit.

\subsection{Architecture comparison}
From the maximum stiffness expressions in Table \ref{port_dyn}, the DOB architectures allow improved rejection of disturbance torques $d$ (i.e. friction in backdriveability) when $|Q|\rightarrow 1$ (i.e. at low frequencies). This allows the DOB to achieve the same steady-state stiffness accuracy with a lower $C$ gain.

In the Z-region analysis, note that $\mathtt{Re}(M(C+C_{ff})I_u)>0$ for PD torque control and typical motor/impedance models. Thus, for a fixed $I_u$, increasing $|C|$ increases the Z-region. For the DOBt, if $\mathtt{Re}(M(C+C_{ff}I_u(1-Q)^{-1})>0$ (met on system parameters here), as $Q\rightarrow1$, the Z-region increases. Other work has shown increasing $|C|$ decreases Z-width \cite{focchi2016}, the Z-region increases with $|C|$, for these three control architectures. 

\subsection{Feedback control}
\begin{figure}
	\centering
	\includegraphics[width=.4\columnwidth]{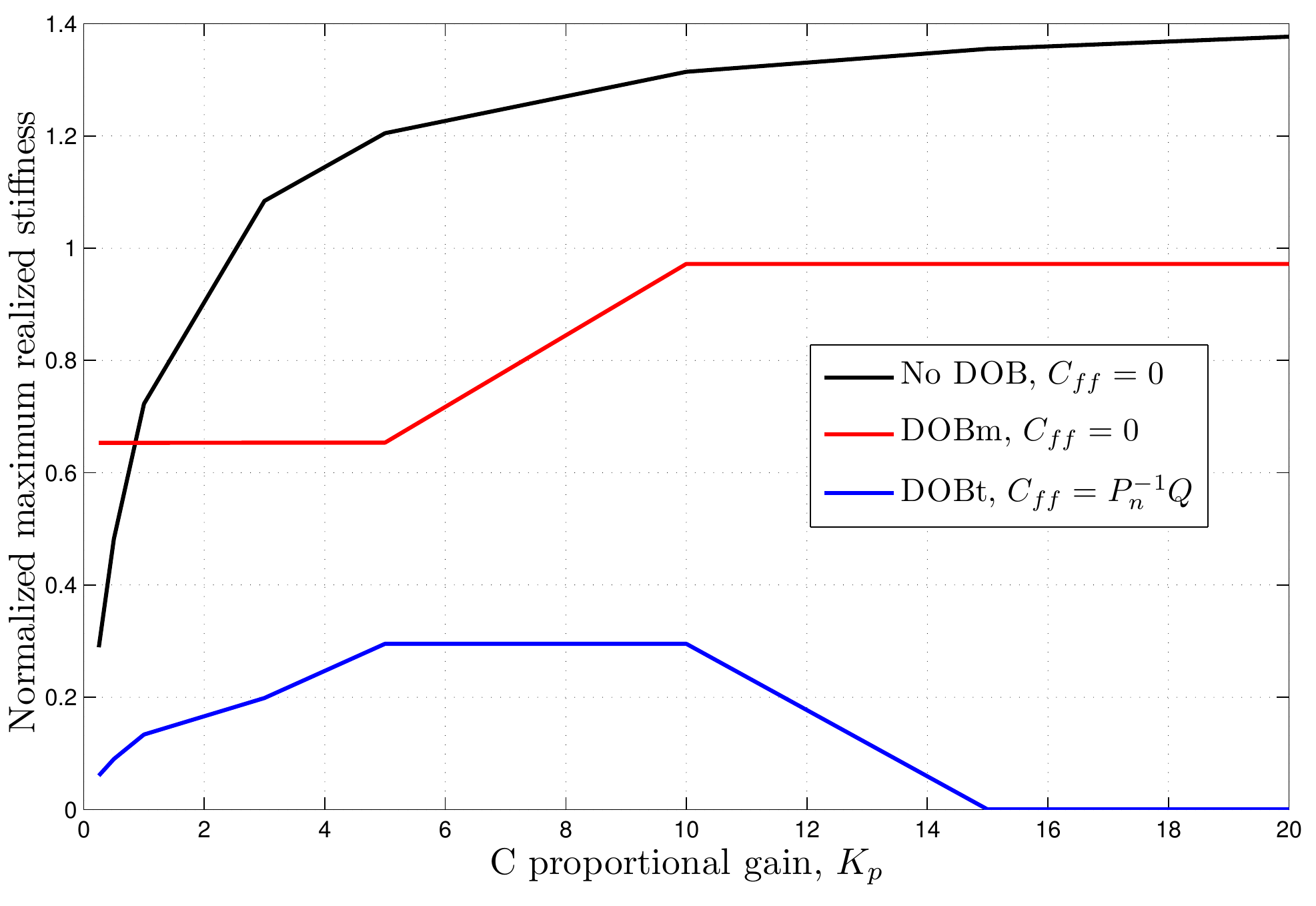} 
	\caption{The maximum stiffness, defined by \eqref{load_pass} with $C_{ff}=0$, which can be rendered under the three architectures, as dependent on (a) the DOB cutoff frequency, and (b) control gain $K_p$.\label{max_stiff_no_cff}}
\end{figure}

Here, the performance of the $\mathrm{DOB_{m}}$ and $\mathrm{DOB_{t}}$ are compared to the no DOB case. Feedforward control $C_{ff}$ is set to the feedforward proposed in each of the original implementations, $C_{ff}= 0$ and $C_{ff}=P_t^{-1}Q$, respectively. The impact of $K_p$ on maximum rendered stiffness can be seen in Figure \ref{max_stiff_no_cff}, showing that while it improves maximum stiffness for the no DOB and DOBm, the DOBt benefits more from a moderate $K_p$ value. For other parameters, the results are monotonic: increasing the $Q$ filter cutoff frequency and $K_d$. Increasing $B_{imp}$ does not make a large impact, and is thus left at $B_{imp}=0$ to reduce noise.
 
\subsection{Feedforward control}
To motivate an improved $C_{ff}$ for the DOBm and DOBt, compensators are proposed to match terms in the torque port dynamics, most easily seen in the rendered stiffness in Table \ref{port_dyn}.
\begin{eqnarray}
C_{ff} = 
\begin{cases}
N(1-Q) & \mathrm{DOBm} \\
P_t^{-1}Q+N(1-Q) & \mathrm{DOBt} \label{cff_good}
\end{cases}
\end{eqnarray}
When the steady-state gain of $Q$ is slightly less than one, $1-Q$ acts as a lead filter, advancing the phase of the system and compensating some of the phase lag. 

\begin{figure}
	\centering
	\includegraphics[width=.5\columnwidth]{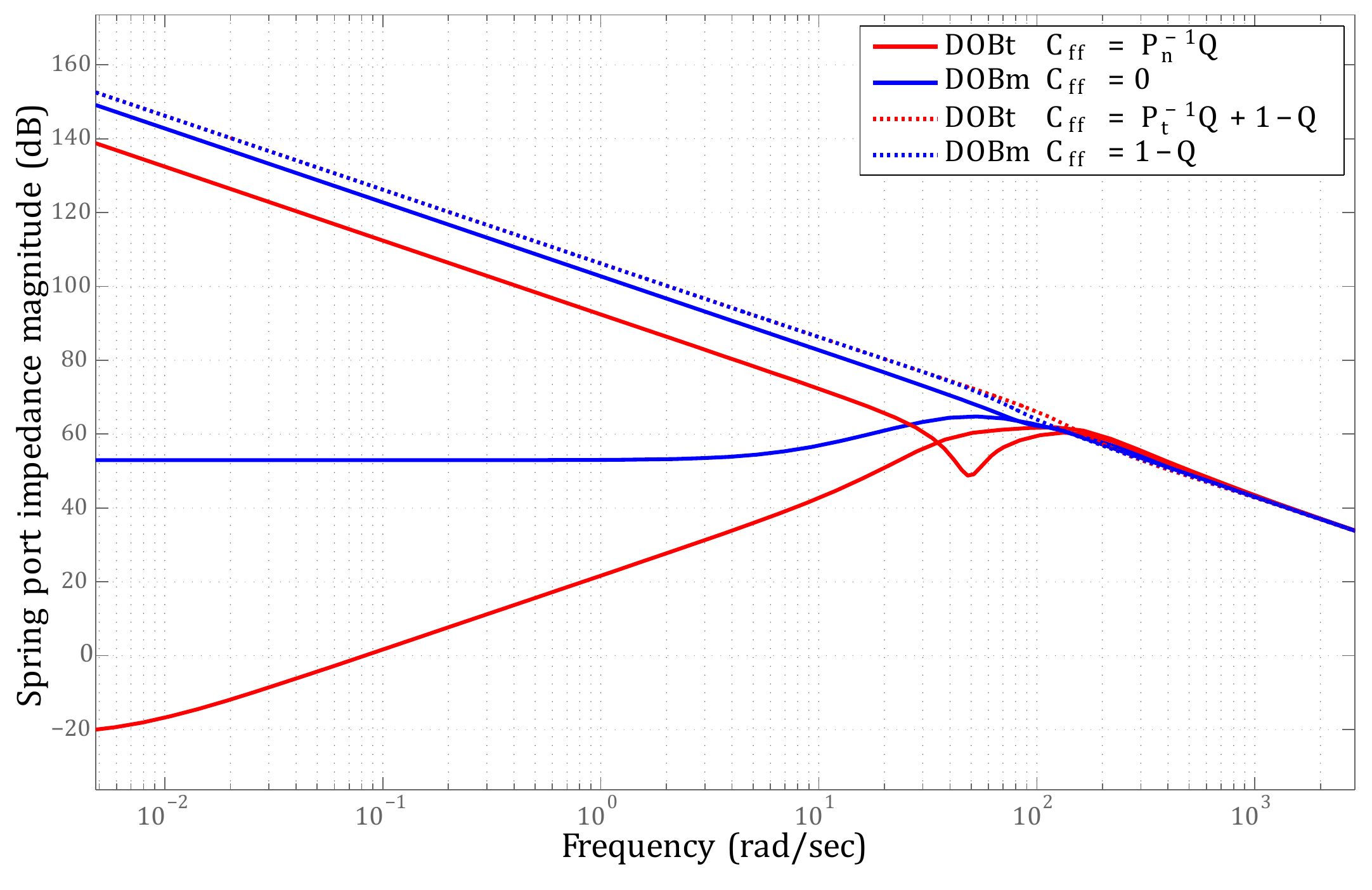}
	\caption{Z-region of DOBm and DOBt under the $C_{ff}$ in \eqref{cff_good}. The advanced feedforward compensation is especially important for the DOBt. Note the low-impedance performance for DOBm and DOBt is unchanged by feedforward control. \label{z-region_cff_comparison}}
\end{figure}


\section{Experimental Validation}
\subsection{Experimental setup}
\begin{figure}[t]
	\centering
	\includegraphics[width=.5\columnwidth]{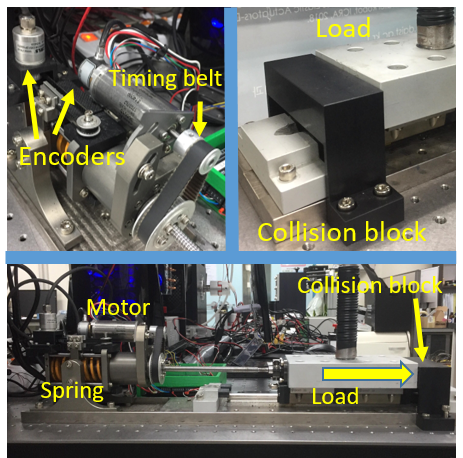}
	\caption{Experimental setup used in validating the real-world coupled stability of the SEA.\label{exp_setup}} 
\end{figure}

For the evaluation and verification of the conditions considering coupled stability at different SEA interaction ports for torque and impedance controllers with DOBm and DOBt, a reaction force series elastic actuator is used (RFSEA) as shown in Figure \ref{exp_setup}. It consists of Brushless DC motor (Maxon EC 4-pole 305015) with a maximum continuous torque of 92.9mNm. The motor  is equipped with an incremental encoder before the gearbox, while the spring deflection is obtained by the high-resolution rotary incremental encoder. 
The physical parameters of the RFSEA are identified by using the FFT analyzer (ONO-SOKKI, CF-9400). In various load conditions, the sweep sine signals were applied to the RFSEA, and the motor velocity and the spring deformation were measured using encoders. Then, the parameters shown in Table \ref{sys_params} were identified.

\subsection{Coupled stability validation}
In order to investigate the conservatism of spring port passivity, and practicality of load port passivity, experiments of the RFSEA tracking a step input to the impedance controller in contact with a two high-stiffness environments were performed. Controller parameters were varied in the ranges $K_{p}=\left[1,\dots10\right]$,$K_{d}=\left[0.01,\dots,.1\right]$, $\omega_{dob} (hz)=\left[10,\dots,100\right]$ under an outer-loop impedance of $K_{imp}=1000$ and $B_{imp}=0.1$.
A binary coupled stability outcome is considered, with an experiment repeated if marginal stability resulted. Trials with sustained or growing oscillation, or a diverging load position, which may result in motor failure, are marked as unstable cases, otherwise, coupled stable. Also, the passivity observer is utilized to identify any passivity violations at both the spring and load ports.

The experimental results are shown in Table \ref{coupled_stab_table}, where the load port significantly reduces the conservatism. This supports using load port passivity as a practical design condition. 
\begin{table}[H]
	\renewcommand{\arraystretch}{1.4}
	\begin{centering}
		\begin{tabular}{|r|c|c|c|c|}
			\hline 
			& \multicolumn{1}{c}{DOBm} &  & \multicolumn{1}{c}{DOBt} & \tabularnewline
			\hline 
			  & \# FP & \# FN & \# FP & \# FN\tabularnewline
			\hline 
			\hline 
			Spring Port Passivity & 0 & 12 & 0 & 15\tabularnewline
			\hline 
			Load Port Passivity & 0 & 0 & 1 & 0\tabularnewline
			\hline 
		\end{tabular}
		\par\end{centering}
	\caption{Experiments are conducted over eighty different controller settings
		and environment stiffnesses, then passivity at the spring and load
		port checked. FP = False Positive, condition true but system not stable
		(ideally 0). FN = False Negative, condition false but system still
		stable (conservatism, ideally 0). In 12 and 15 cases, the system was
		not passive at the spring port, but achieved coupled stability experimentally. \label{coupled_stab_table}}
\end{table}

\subsection{Coupled stability discussion} 
The objective of the experiment is to validate the conditions developed for coupled stability at different interaction ports. For both DOBm and DOBt, the torque and impedance PD controller gains and DOB bandwidth are varied to check coupled stability. A binary coupled stability condition is considered without consideration of tracking performance.
At first, the conditions on the inner torque control loop are evaluated with two coupling conditions: the load is fixed at one side by a metallic block and load collision to a metallic block. The white arrow in Figure \ref{exp_setup} shows the direction of the collision and reference direction for the fixed load condition. In torque-control mode, higher PD controller gains typically result in hardware failure  rather than instability. Thus, for torque control, it is difficult to obtain an unstable case even in impact scenarios, and coupled stability cannot be evaluated in torque control on this experimental setup.
For impedance control, instability occured, and the conditions can be evaluated appropriately. For DOBm, the coupled stability condition is insensitive to higher values of $K_{imp}$, unlike the other controller gains. High $K_p$ and $B_{imp}$ can cause instabilty. For instance, experiment results show that decreasing the value of $K_{imp}$ cannot restore couple stability from high $K_p$ and $B_{imp}$. Whereas, DOBt allows higher gain values than DOBm with regards to inner torque control; however, it is very sensitive to higher values of $B_{imp}$ and $K_d$. For instance, unlike in DOBm, $B_{imp}$ can go as high as 50, but for DOBt, it is limited to values less than 1 on this experimental setup.

\subsection{Maximum rendered stiffness validation}
Experimentally determining the maximum virtual stiffness $K_{imp}$ that can be rendered safely is done by finding the maximum stiffness where load port passivity is held for the No DOB, DOBm and DOBt architectures. 
\begin{table}[H]
	\renewcommand{\arraystretch}{1.4}
	\begin{centering}
		\begin{tabular}{|r|c|c|c|c|}
			\hline 
			& Spring pass. & Load pass. & Simulation  & Experiment\tabularnewline
			\hline 
			\hline 
			No DOB & $1$ & $1.93$ & $1.93$ & $\sim1.7$\tabularnewline
			\hline 
			DOBm & $1$ & $1.93$ & $1.93$ & $\sim1.8$\tabularnewline
			\hline 
			DOBt & $1$ & $1.85$ & $1.85$ & $\sim1.65$\tabularnewline
			\hline 
		\end{tabular}
		\par\end{centering}
	\caption{The maximum normalized stiffness achieved from analytical consideration of passivity, simulation, and experimental results. The DOBm allows for a slightly higher high stiffness than the DOBt and no DOB.\label{stiff_results}}
	\end{table}

Table \ref{stiff_results} supports the following points: (i) $K_{imp}$ can be safely set higher than $K$; (ii) load port condition can give a less conservative and more practical limit for $K_{imp}$, in both simulation and experiments. The difference between the simulation and experimental results could be caused by a number of reason such as load side uncertainties, time-delay, or modelling the 3-mass system as a 2-mass.

\begin{figure}[t]
	\centering
	\includegraphics[width=.5\columnwidth]{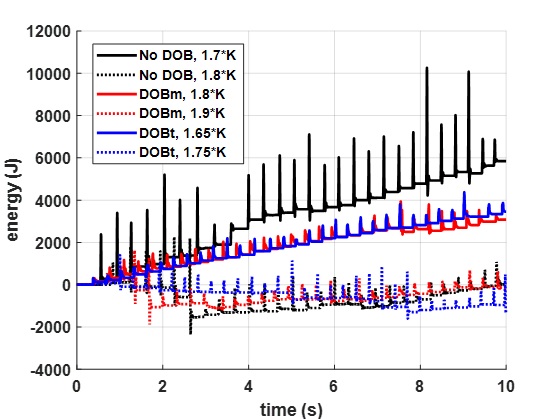}
	\caption{Passivity at the load-port for high stiffnesses  The stiffness boundary is validated as passivity is lost between the higher and lower stiffness on each architecture.\label{PO_results}} 
\end{figure}

In these experiments, a high-stiffness hammer is used to strike the static load side, where a load cell was installed to measure $\tau_{ext}$. This allows evaluation of load port passivity using passivity observer, shown in Figure \ref{PO_results} where decreasing energy is a passivity violation, and thus not safe according to the load port condition. 

\begin{figure}[t]
	\centering
	\includegraphics[width=.5\columnwidth]{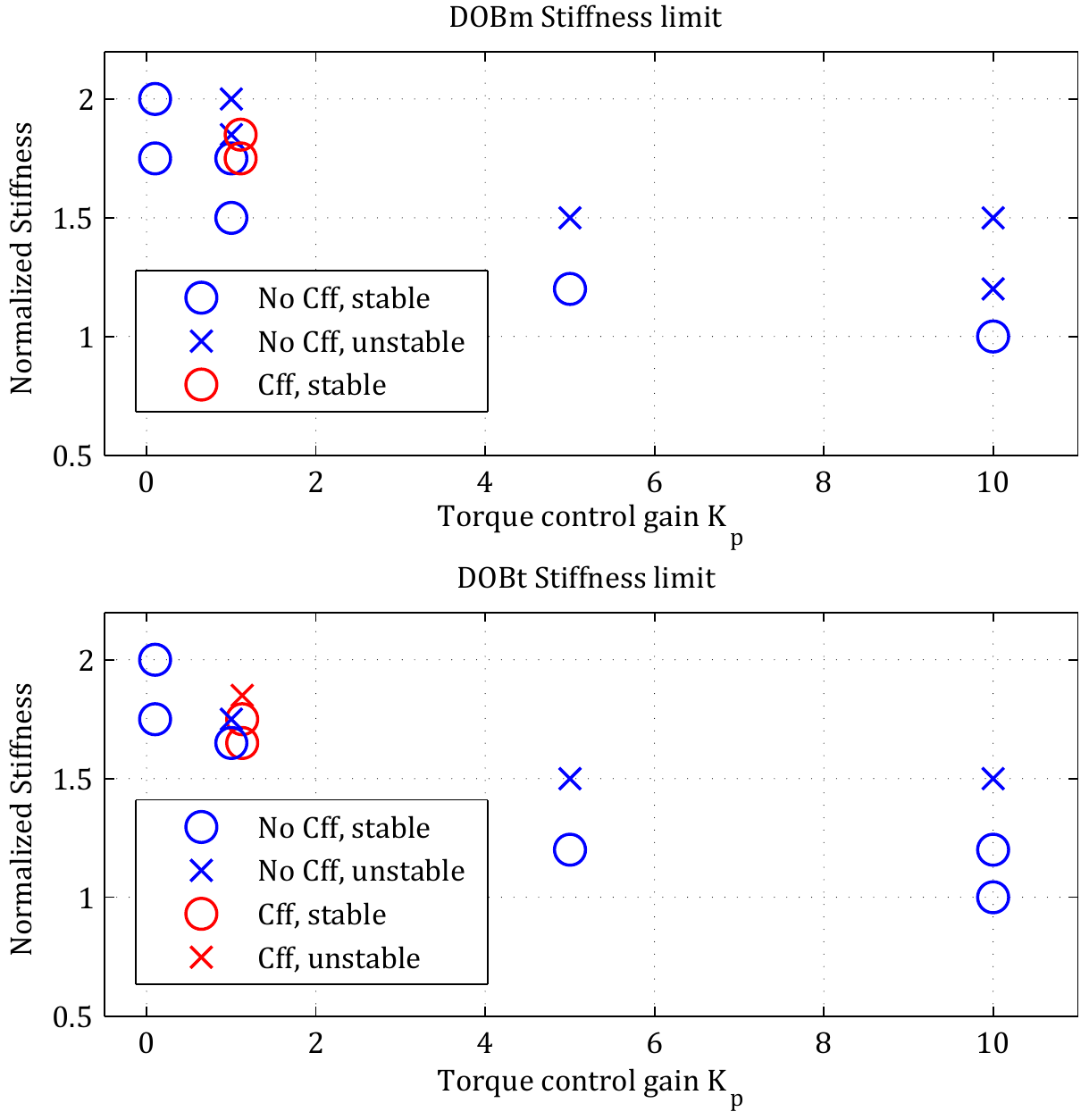}
	\caption{Maximum rendered stiffness over $K_p$, with and without $C_{ff}$.  $K_d = .01$, $B_{imp}=0$, and $\omega_{DOB}=25$ Hz for all. \label{max_stiff_compare}} 
\end{figure}

\begin{table}
	\centering
	\renewcommand{\arraystretch}{1.2}
	\begin{tabular}{rl rl}
		\hline 
		\hline
		$J_{m}$ & $6.4e-6$ $\mathrm{kg\cdot m^{2}}$ & $J_{l}$ & $7$ $\mathrm{kg\cdot m^{2}}$\\ 
		$B_{m}$ & $6e-5$ $\mathrm{Nm\cdot s/rad}$ & $B_{l}$ & $100$ $\mathrm{Nm\cdot s/rad}$\\
		$K$ & $141350$ $\mathrm{N/m}$ & $N$ & $1/7854$ $\mathrm{rad/m}$ \\
		\hline 
		\hline
	\end{tabular}
	\caption{Identified Parameters\label{sys_params}}
\end{table}
Further experiments are undertaken to validate the impact of $K_p$ and $C_{ff}$ on the maximum rendered stiffness.  In Figure \ref{max_stiff_compare}, the stable and unstable responses can be seen (x and o, respectively) with and without the $C_{ff}$ in \eqref{cff_good}. As $K_p$ increases, maximum $K_{imp}$ decreases, although the very low gain ($K_p=.1$) may not render the impedance accurately.  The proposed $C_{ff}$ increases the maximum safe stiffness on both the DOBm and DOBt, although the contribution is more minor than suggested analytically. 

\section{Conclusions}
This paper has established initial results analyzing DOB control of SEAs from the perspective of passivity, towards rendering high impedance in a high-stiffness environment. While passivity does not perfectly correspond to real-world coupled stability, using the load dynamics allows for a less-conservative result which can help guide controller design. The use of a DOB reduces the need for high-gain torque feedback, and can be improved with the use of dynamic feedforward compensators. 

\bibliographystyle{ieeetr}
\bibliography{lib}
\end{document}